%% file: main.tex
\documentclass[10pt,twocolumn,letterpaper]{article}

\usepackage{cvpr}              
\definecolor{darkgreen}{RGB}{1,50,32}
\definecolor{forestgreen}{RGB}{34,139,34}
\usepackage{pifont}

\usepackage{multirow}
\usepackage{ctable}


\definecolor{cvprblue}{rgb}{0.21,0.49,0.74}
\usepackage[pagebackref,breaklinks,colorlinks,allcolors=cvprblue]{hyperref}
\usepackage{pifont}
\usepackage{commath}

\usepackage{multirow}
\usepackage{ctable}
\def\paperID{13} 
\def\confName{CVPR}
\def\confYear{2025}

\title{OV-COAST: Cost Aggregation with Optimal Transport for Open-Vocabulary Semantic Segmentation}

\author{ Aditya Gandhamal*$\ ^{1}$, Aniruddh Sikdar*$\ ^{2}$, Suresh Sundaram$ ^{2,3}$        
\\  
$^{1}$Kotak IISc AI-ML Centre, Indian Institute of Science, Bengaluru, India \\
$^{2}$Robert Bosch Centre for Cyber Physical Systems, Indian Institute of Science, Bengaluru, India\\
$^{3}$Department of Aerospace Engineering, Indian Institute of Science, Bengaluru, India\\
}

\begin{document}
\maketitle
\input{sec/0_abstract}    
\input{sec/1_intro}

\input{sec/2_formatting}

\input{sec/3_finalcopy}
{
    \small
    \bibliographystyle{ieeenat_fullname}
    \bibliography{main}
}


\end{document}

%% file: sec/0_abstract.tex
\begin{abstract}

Open-vocabulary semantic segmentation (OVSS) entails assigning semantic labels to each pixel in an image using textual descriptions, typically leveraging world models such as CLIP. To enhance out-of-domain generalization, we propose Cost Aggregation with Optimal Transport (OV-COAST) for open-vocabulary semantic segmentation. To align visual-language features within the framework of optimal transport theory, we employ cost volume to construct a cost matrix, which quantifies the distance between two distributions. Our approach adopts a two-stage optimization strategy: in the first stage, the optimal transport problem is solved using cost volume via Sinkhorn distance to obtain an alignment solution; in the second stage, this solution is used to guide the training of the CAT-Seg model.
We evaluate state-of-the-art OVSS models on the MESS benchmark, where our approach notably improves the performance of the cost-aggregation model CAT-Seg with ViT-B backbone, achieving superior results, surpassing CAT-Seg by 1.72\% and SAN-B by 4.9\%  mIoU. The code is available at \href{https://github.com/adityagandhamal/OV-COAST/}{https://github.com/adityagandhamal/OV-COAST/}
\end{abstract}

%% file: sec/1_intro.tex
\section{Introduction}
\label{sec:intro}


Advancing deep learning models with strong cross-domain generalization is essential for real-world computer vision applications \cite{sikdar2022fully, sikdar2023deepmao, udupa2024mrfp, sikdar2023fully, sikdar2025ogp}.  Open-vocabulary semantic segmentation (OVSS) addresses this need by enabling pixel-level classification over open category sets through the use of natural language as semantic guidance. Recent large-scale vision-language models, pre-trained on vast internet-scale data with natural language supervision, demonstrate strong generalization capabilities \cite{shi2024lca, radford2021learning, jia2021scaling, yuan2021florence}. Despite being trained with image-text pairs, these models depend on coarse image-level supervision, which limits their accuracy and effectiveness in fine-grained pixel-level segmentation tasks \cite{zhou2022extract, ding2022decoupling}.

\begin{figure}[t]
    \centering
    \includegraphics[scale=0.19]{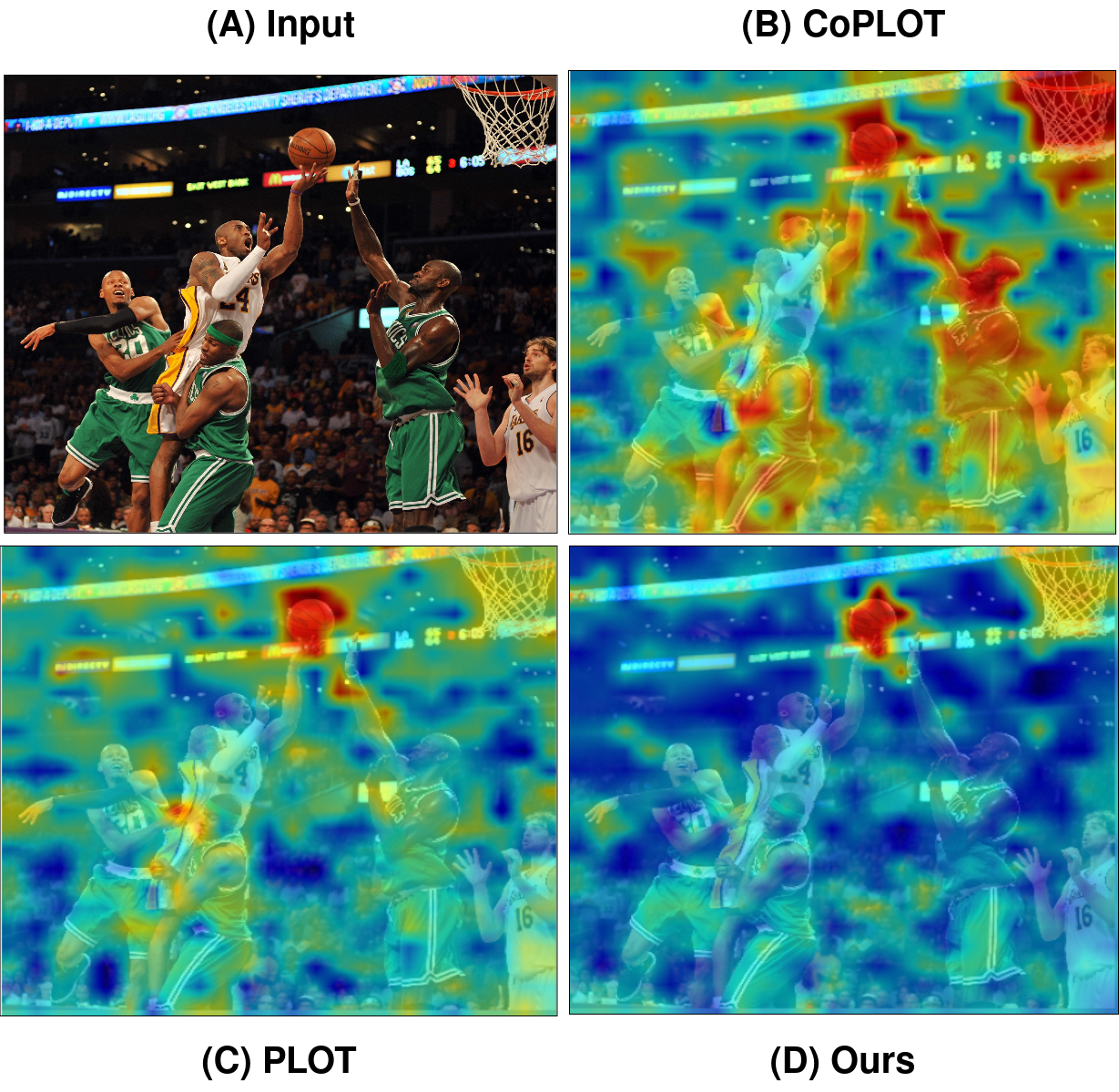}
    \setlength{\abovecaptionskip}{-5pt} 
    \setlength{\belowcaptionskip}{-11pt}
    \caption{Visualization of logits of the vision and text embeddings from the ViT-B CLIP encoder. The heatmaps correspond to the prompt learning techniques applied to CAT-Seg with optimal transport: (a) Input image, (b) CoPLOT, (c) PLOT, (d) Ours. The highlighted regions indicate the areas most correlated with the text description: ``A photo of a ball in the scene."}
    \label{fig:grad_plot}
\end{figure}

\begin{figure}[t]
    \centering
    \includegraphics[scale=0.325]{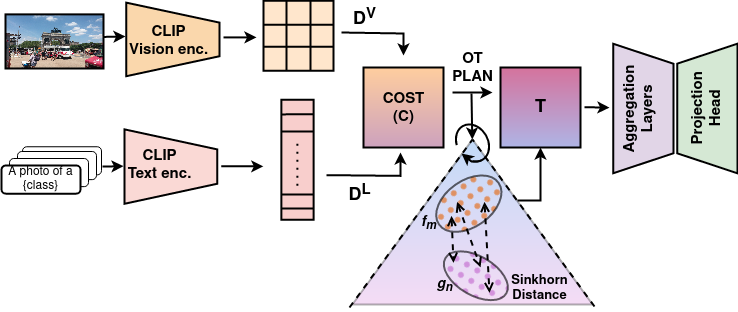}
    \caption{Overview of OV-COAST. We compute cost from image embeddings $D^{V}$ and text embeddings $D^{L}$, obtained from the ViT-B CLIP encoders. We use the Sinkhorn Distance to compute the optimal transport plan to match the text features with the image features.} 
    \label{fig:net-architecture}
\end{figure}

Recent studies have improved these models for segmentation tasks through innovative approaches, including (1) mask proposal networks  \cite{ghiasi2022scaling, xu2022simple, xu2023side, yu2023convolutions} and (2) using cost aggregation methods \cite{cho2024cat}.    However, these models face challenges in pixel-level classification for unseen classes, particularly when exposed to significant domain shifts \cite{blumenstiel2024mess}. This challenge intensifies as the number of unseen categories increases \cite{ding2022decoupling}, with open-vocabulary models becoming more prone to overfitting during adaptation to downstream tasks \cite{wortsman2022robust,kumar2022fine}. While OVSS models are effective at capturing open-set visual concepts for segmentation, their zero-shot generalization performance significantly declines when encountering substantial shifts in categories, distributions, or domains between large-scale pretraining datasets and downstream tasks \cite{zhang2024dept}. This constrains the practical deployment of open-vocabulary models in real-world scenarios.\\
To address this, we introduce Cost Aggregation with Optimal Transport (OV-COAST) for Open-Vocabulary Semantic Segmentation. CAT-Seg \cite{cho2024cat} combines visual and textual embeddings from CLIP encoders \cite{radford2021learning} to create a multi-modal cost aggregation based on cosine similarity \cite{liu2022graftnet, rocco2017convolutional}. 
To align the features using optimal transport theory  \cite{monge1781memoire}, we apply the cost volume to construct a cost matrix, which is then used to compute the distance between two distributions through multiple sampling. 
We adopt a two-stage optimization strategy \cite{chen2022plot}. In the first stage, the inner loop keeps both visual and textual features fixed while solving the optimal transport problem via Sinkhorn distance by using cost volume. In the second stage, the outer loop freezes the optimal transport parameters and trains the CAT-Seg model for the segmentation task. Table \ref{table:cost-op} provides a comparison of existing architectures and outlines the differences with OV-COAST.  OV-COAST performs better than other techniques using optimal transport for OVSS, illustrated by the Grad-CAM images presented in Fig. \ref{fig:grad_plot}.
We conduct extensive evaluations of OV-COAST on the MESS benchmark, using 19 datasets from diverse domains for semantic segmentation. Our approach improves the performance of the cost-aggregation model, CAT-Seg \cite{cho2024cat}, achieving superior results on the MESS benchmark.

Our key contributions are as follows: (1) We propose OV-COAST, a method that leverages optimal transport and cost aggregation to achieve improved multi-modal alignment in the CLIP embedding space, supported by a two-stage optimization strategy. (2) Extensive experiments validate the superiority of OV-COAST, achieving state-of-the-art performance on the MESS benchmark \cite{blumenstiel2024mess}.

\begin{table}[]
\centering
\begin{tabular}{l|cc|c}
\hline
\multicolumn{1}{c|}{\textbf{Method}} & \textbf{MPG} & \textbf{MTP} & \textbf{ITM}                                                                    \\ \hline
ZegFormer \cite{ding2022decoupling}                                            & \ding{51}            & \ding{55}            & -                                                                               \\
ZSseg \cite{xu2022simple}                                                        & \ding{51}            & \ding{55}            & -                                                                               \\
ZegCLIP \cite{zhou2022zegclip}                                                      & \ding{55}            & \ding{51}            & MHA                                                                             \\
OVSeg \cite{liang2023open}                                                        & \ding{51}            & \ding{51}            & Contrastive Loss                                                                \\
SAN \cite{xu2023side}                                                          & \ding{51}            & \ding{55}            & -                                                                               \\
CAT-Seg \cite{cho2024cat}                                                      & \ding{55}            & \ding{51}            & Cost Aggregation                                                                \\ \hline
OV-COAST (ours)                                                         & \ding{55}            & \ding{51}            & \begin{tabular}[c]{@{}c@{}}Cost Aggregation + \\ Optimal Transport\end{tabular} \\ \hline
\end{tabular}
\caption{Comparison of ViT-based CLIP methods for OVSS, incorporating mask proposal generators (MPG), multiple text prompts (MTP), and image-text matching (ITM). MHA stands for Multihead Attention.}
\label{table:cost-op}
\end{table}

\begin{table*}[]
\resizebox{1.02\textwidth}{!}{%
\begin{small}
\begin{tabular}{ccccccccccccccccccccc}
\specialrule{.1em}{.05em}{.05em} 
\textbf{}       & \rotatebox{90}{\textbf{Dark Zurich}} & \rotatebox{90}{\textbf{MHP v1}} & \rotatebox{90}{\textbf{FoodSeg103}} & \rotatebox{90}{\textbf{ATLANTIS}} & \rotatebox{90}{\textbf{iSAID}} & \rotatebox{90}{\textbf{ISPRS}} & \rotatebox{90}{\textbf{WorldFloods}} & \rotatebox{90}{\textbf{FloodNet}} & \rotatebox{90}{\textbf{UAVid}} & \rotatebox{90}{\textbf{Kvasir-inst.}} & \rotatebox{90}{\textbf{CHASE DB1}} & \rotatebox{90}{\textbf{CryoNuSeg}} & \rotatebox{90}{\textbf{Corrosion CS}} & \rotatebox{90}{\textbf{DeepCrack}} & \rotatebox{90}{\textbf{PST900}} & \rotatebox{90}{\textbf{ZeroWaste-f}} & \rotatebox{90}{\textbf{SUIM}} & \rotatebox{90}{\textbf{CUB-200}} & \rotatebox{90}{\textbf{CWFID}} & \rotatebox{90}{\textbf{Mean}} \\ \specialrule{.1em}{.05em}{.05em} 
Random (L.B)    & 1.31                 & 1.27            & 0.23                & 0.56              & 0.56           & 8.02           & 18.43                & 3.39              & 5.18           & 27.99                 & 27.25              & 31.25              & 9.3                   & 26.52              & 4.52            & 6.49                 & 5.3           & 0.06             & 13.08          & 10.27         \\
Best Sup. (U.B) & 63.9                 & 50.0            & 45.1                & 42.22             & 65.3           & 87.56          & 92.71                & 82.22             & 67.8           & 93.7                  & 97.05              & 73.45              & 49.92                 & 85.9               & 82.3            & 52.5                 & 74.0          & 84.6             & 87.23          & 70.99         \\
ZSSeg-B         & 16.86                & 7.08            & 8.17                & 22.19             & 3.8            & 11.57          & 23.25                & 20.98             & 30.27          & 46.93                 & 37.0               & \textbf{38.7}               & 3.06                  & 25.39              & 18.76           & 8.78                 & 30.16         & 4.35             & 32.46          & 22.73         \\
ZegFormer-B     & 4.52                 & 4.33            & 10.01               & 18.98             & 2.68           & 14.04          & 25.93                & 22.74             & 20.84          & 27.39                 & 12.47              & 11.94              & 4.78                  & 29.77              & 19.63           & 17.52                & 28.28         & 16.8             & 32.24          & 17.57         \\
X-Decoder-T     & 24.16                & 3.54            & 2.61                & 27.51             & 2.43           & 31.47          & 26.23                & 8.83              & 25.65          & 55.77                 & 10.16              & 11.94              & 1.72                  & 24.65              & 19.44           & 15.44                & 24.75         & 0.51             & 29.25          & 19.8          \\
SAN-B           & 24.35                & 8.87            & 19.27               & 36.51             & 4.77           & 37.56          & 31.75                & 37.44             & 41.65          & \textbf{69.88}                 & 17.85              & 11.95              & 19.73                 & \underline{50.27}              & 19.67           & \textbf{21.27}                & 22.64         & \textbf{16.91}            & 5.67           & 26.21         \\
OpenSeeD-T      & 28.13                & 2.06            & 9.0                 & 18.55             & 1.45           & 31.07          & 30.11                & 23.14             & 39.78          & \underline{59.69}                 & \textbf{46.88}              & 33.76              & 13.38                 & 47.84              & 2.5             & 2.28                 & 19.45         & 0.13             & 11.478         & 21.82         \\ 
Gr.-SAM-B       & 20.91                & \textbf{29.38}           & 10.48               & 17.33             & 12.22          & 26.68          & 33.41                & 19.19             & 38.35          & 46.82                 & 23.56              & \underline{38.06}              & \textbf{20.88}                 & \textbf{59.02}              & \textbf{21.39}           & 16.74                & 14.13         & 0.43             & 38.41          & 25.65         \\
ODISE           & \underline{28.88}                & 7.31            & 16.62               & 36.73             & 11.71          & \textbf{54.83}          & 26.30                & 35.06             & \textbf{43.46}          & 28.02                 & 20.60              & 12.27              & 10.74                 & 24.34              & \underline{20.18}           & 5.74                 & 26.24         & 4.62             & 30.43          & 23.37         \\
CAT-Seg         & 28.86                & 23.74           & \textbf{26.69}               & \textbf{40.31}             & \underline{19.34}          & 45.36          & \underline{35.72}                & \underline{37.57}             & 41.55          & 48.2                  & 16.99              & 15.7               & 12.29                 & 31.67              & 19.88           & 17.52                & \textbf{44.71}         & 10.23            & \underline{42.77}          & \underline{29.42}         \\ \hline
OV-COAST        & \textbf{28.94}               & \underline{25.15}           & \underline{25.92}               & \underline{39.59}            & \textbf{19.87}         & \underline{46.02}          & \textbf{37.22}                & \textbf{37.59}             & \underline{43.16}          & 46.56                 & \underline{30.30}             & 15.76              & \underline{14.47}                 & 44.22              & 19.66           & \underline{18.0}                & \underline{44.61}        & \underline{10.52}           & \textbf{44.34}         & \textbf{31.15}      \\ \specialrule{.1em}{.05em}{.05em} 
\end{tabular}
\end{small}
}
\caption{Quantitative evaluation on the MESS benchmark, which encompasses datasets from various domains, presents substantial domain shifts for OVSS models. The 'Random' metric represents predictions distributed uniformly, whereas 'Best Supervised' (\textit{Best Sup.}) denotes the performance upper bound for each dataset. All models use a ViT-B backbone or similar architecture to ensure a fair comparison.}
\label{mess}
\end{table*}

\section{Related Works}

\textbf{Open-vocabulary semantic segmentation (OVSS)} Large-scale pre-training on image-text pairs enables joint visual-textual representation in a shared semantic space. Open-vocabulary semantic segmentation (OVSS) aims to bridge the gap between image-level and pixel-level understanding by enabling segmentation of unseen classes. OVSS models are generally categorized into two main types: (i) mask-proposal networks \cite{ghiasi2022scaling, ding2022open, xu2023open, xu2022simple, xu2023side, yu2023convolutions}, and (ii) cost-aggregation networks \cite{cho2024cat}. Mask-proposal approaches generate class-agnostic region proposals, which are then semantically aligned with CLIP-derived textual embeddings to facilitate open-vocabulary classification at the pixel level. For instance, OpenSeg \cite{ghiasi2022scaling} leverages local image regions to form proposals, whereas models like ZegFormer \cite{ding2022decoupling} and ZSseg \cite{xu2022simple} adopt a two-stage architecture to address this task.   CAT-Seg \cite{cho2024cat} enhances robustness by using similarity scores to combine CLIP’s image and text embeddings into a cost volume, rather than relying on direct embeddings.  OVSS models are often evaluated on in-domain datasets \cite{zhou2019semantic,everingham2015pascal},  which fail to reflect the domain shifts and environmental diversity of real-world scenarios. Cost-aggregation networks have shown enhanced resilience to domain shifts compared to other architectures, highlighting the need to bridge this gap for effective real-world deployment of OVSS models. \\
\textbf{Optimal Transport}  Optimal Transport \cite{monge1781memoire} was initially introduced to minimize the cost of moving multiple items at once and is now widely used to compare distributions represented as feature sets \cite{peyre2019computational}. PLOT \cite{chen2022plot} uses optimal transport (OT) to align visual features with textual prompts, enabling fine-grained matching between the two modalities. PLOT demonstrates how the performance of CoOp \cite{zhou2022learning} can be enhanced using optimal transport (OT), where learnable prompts enable a flexible alignment between visual and textual modalities through an adaptive transport plan. Optimal transport (OT) is still in its early stages of application for segmentation tasks. ZegOT \cite{kim2023zegot} uses OT to align textual prompts with frozen image embeddings, bridging the gap between pretrained VLM features and learnable prompts.



%% file: sec/2_formatting.tex
\begin{table}[]
\centering
\begin{tabular}{c|c}
\hline
\textbf{Method}                   & \textbf{\begin{tabular}[c]{@{}c@{}}mIoU\\ (on MESS)\end{tabular}} \\ \hline
CAT-Seg + VPT + Optimal Transport & 20.68                                                             \\
CAT-Seg + CoPLOT                  & 23.98                                                             \\
CAT-Seg + PLOT                    & 26.61                                                             \\ \hline
OV-COAST                     & \textbf{31.15}                                                    \\ \hline
\end{tabular}
\caption{Performance comparison of prompt learning techniques with optimal transport applied to CAT-Seg on the MESS benchmark.}
\label{table:prompt-ot}
\end{table}

\begin{figure*}[t]
    \centering
    \includegraphics[scale=0.165]{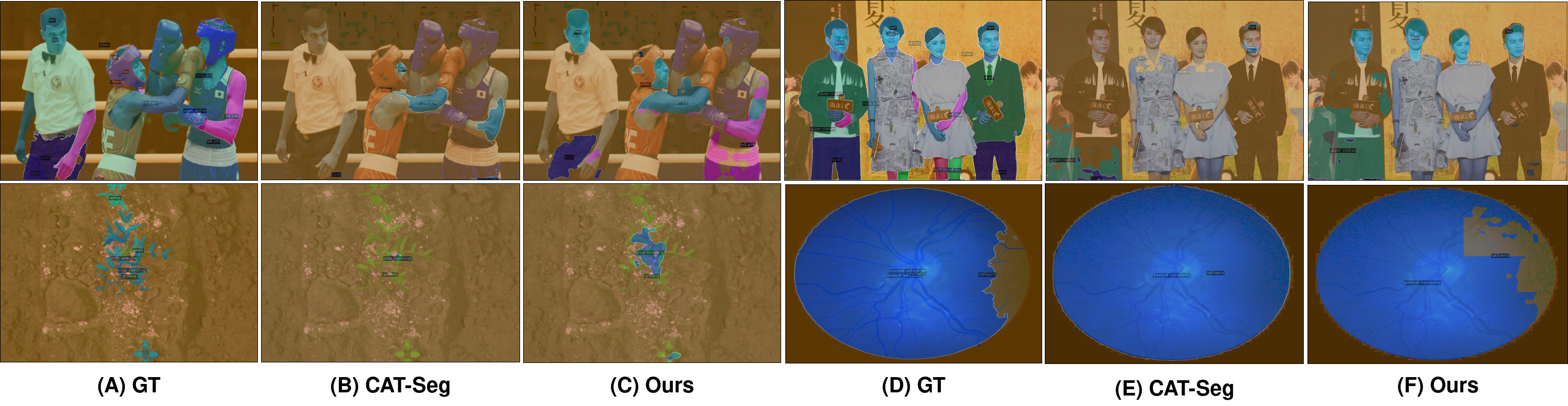}
    \caption{Qualitative comparison of CAT-Seg and OV-COAST(ours) on MHP, CWFID and CHASE datasets from the MESS benchmark.}
    \label{fig:radar_plot}
\end{figure*}

\section{Preliminaries}
We provide a brief overview of Optimal Transport \cite{monge1781memoire} and the PLOT framework \cite{chen2022plot}. The discrete distributions of the feature representations are defined as follows:
\begin{equation}
        U = \sum_{m=1}^{M} u_m \delta_{D^{V}} \quad and \quad V = \sum_{n=1}^{N}  v_n \delta_{D^{L}}
\end{equation}
here, \textit{u} and \textit{v} are the discrete probability vectors, $\delta$ denotes the Dirac delta function, $D^{V}$ represents the visual features from the CLIP backbone, and $D^{L}$ represents the textual features. The overall distance is given by,
\begin{equation}
    <T,C> = \sum_{m=1}^{M} \sum_{n=1}^{N} T_{m,n}C_{m,n}
\end{equation}
where $C$ is the cost matrix, with $C_{m,n}$ = 1- sim ($D^{V}$, $D^{L}$) representing the cost between $D^{V}$ and $D^{L}$. \textit{T} denotes the transport plan that minimizes the total distance, leading to the following optimal transport optimization problem:
\begin{align}
  &d_{OT}(u,v|C) =  \arg\min_{T} <T,C> \\
    &\text{subject to} \quad T1_{N} = u, T^{T}1_{M} = v, T \in \mathbb{R}_{+}^{M\times N}
\end{align}

A fast optimization solution can be obtained in a few iterations by leveraging the Sinkhorn distance \cite{cuturi2013sinkhorn}, formulated as follows:

\begin{equation}
    T^* = diag(u^{(t)})exp(-C/\lambda)diag(v^{(t)})
\end{equation}
where $t$ denotes the iteration step, and both $u^{t}$ and $v^{t}$ are initialized following PLOT \cite{chen2022plot}. PLOT incorporates learnable text prompts based on  CoOp \cite{zhou2022learning}. During the inner training loop, the transport plan $T$ is learned by minimizing the OT distance, effectively pushing $D^{L}$ closer to $D^{V}$. Prior to the loop, $\delta$, $u^{t}$, and $v^{t}$ are initialized. The loop terminates once the change in $v^{t}$, denoted as $\Delta_{v}$, falls below a predefined threshold. In the outer loop, the transport plan $T^*$ is kept constant while the model is trained on a classification task using cross-entropy loss, and only the prompts are updated accordingly. For detailed explanation, please refer to \cite{chen2022plot}.

\section{Cost Aggregation with Optimal Transport (OV-COAST)}

\textbf{Problem Formulation} Let $I$ represent an image associated with a set of  classes  $C$ = \{$C_{1}$, $C_{2}$, ..., $C_{N}$\}, where each class $C_{i}$ is defined by a textual label. OVSS models aim to assign a class label $C_{i}$ for each pixel in an image $I$, where the candidate class set $N$ can vary at inference and may include novel categories unseen during fine-tuning on the training data.

\subsection{OV-COAST}

We adopt CAT-Seg \cite{cho2024cat} as our baseline for its strong robustness to domain shifts compared to other models. Building on this, we propose cost aggregation with optimal transport (OV-COAST). For a given image \textit{I} and its corresponding textual descriptions, the CLIP image and text encoders \cite{radford2021learning} are used to obtain image embeddings $D^{V}$ and text embeddings $D^{L}$. The embeddings are then integrated to form a multi-modal cost volume, using cosine similarity \cite{rocco2017convolutional}. The cost volume  is expressed as:

\begin{equation}
C_{aggr}\{i,n\} =\frac{D^{V}(i)\cdot D^{L}(n)}{||D^{V}(i)||\cdot ||D^{L}(n)||} \
\end{equation}

The cost matrix defined in OT, equation (2), is obtained using the cost volume method similar to CAT-Seg. Specifically, we define the cost matrix as $C_{m,n}$ = $(1- C_{aggr}\{i,n\})$. The cost volume embedding is passed to the spatial cost aggregation block for each class, and then input to the aggregation stage, which includes both spatial and class aggregation modules \cite{kendall2017end}, as shown in Fig. \ref{fig:net-architecture}. The output embeddings $F'\{:,n\}$ are then forwarded to the class aggregation module to model the relationships between different class categories. Bilinear sampling is applied to the concatenated aggregated and intermediate feature maps from the CLIP image encoder. This process is repeated twice to generate a high-resolution feature map, which is then passed to the prediction head.

During training, following PLOT, we adopt a two-stage optimization strategy. In the first stage, the inner loop iteratively optimizes the transport plan $T^*$ by minimizing the OT distance, effectively aligning the visual embeddings $D^{V}$ with the textual embeddings $D^{L}$ through the newly formulated cost matrix via Sinkhorn distance.  This process continues until the change in $v^{t}$, denoted as $\Delta_{v}$, falls below a predefined convergence threshold. In the outer loop, the transport plan $T^*$ remains fixed while the model is trained for the OVSS task using cross-entropy loss, following the same training strategy as CAT-Seg. Unlike PLOT, no additional prompting techniques are employed in this phase. Both CLIP encoders utilize attention fine-tuning, similar to the training approach used in CAT-Seg. For more information on the CAT-Seg model, please refer to \cite{cho2024cat}.

\section{Experiments}

\textbf{Datasets and Evaluation}
All models are evaluated using the mean Intersection over Union (mIoU) metric and are trained on the COCO-Stuff dataset  \cite{caesar2018coco}.   We follow the same training strategy as CAT-Seg. ViT-B/16 transformer backbone pre-trained with CLIP is used. In the two-stage optimization process,  $\Delta_{v}$   is set to 0.01, and the inner loop iterates until this convergence threshold is met. The MESS benchmark is used to evaluate out-of-domain performance across 22 diverse datasets, with our model tested on 19 of them. \\
\textbf{Multi-domain OVSS} Table \ref{mess} presents a comparative analysis of OVSS models on the MESS benchmark. As shown, OV-COAST surpasses CAT-Seg by an average of 1.72\% and outperforms SAN-B by 4.94\% in mIoU.
Qualitative results across multiple domains in the MESS benchmark demonstrate the superior predictive performance of OV-COAST, which is further illustrated by the Grad-CAM images presented in Fig. \ref{fig:grad_plot}.\\
\textbf{Ablation study with prompt learning techniques} As shown in Table \ref{table:prompt-ot}, contrary to the improvements seen with PLOT, introducing prompt learning techniques alongside optimal transport in CAT-Seg results in a performance drop, even falling below  CAT-Seg.

\section{Conclusion} 
This paper introduces  Cost Aggregation with Optimal Transport (OV-COAST) for Open-Vocabulary Semantic Segmentation, a framework designed to evaluate and enhance the robustness of OVSS models on out-of-domain datasets. We use CAT-Seg as our baseline, which aggregates costs using cosine similarity between visual and textual embeddings from CLIP. We use the cost volume to construct a cost matrix within the framework of optimal transport theory. In the first stage, the inner loop solves the optimal transport problem using the computed cost volume via Sinkhorn distance. In the second stage, the outer loop freezes the optimal transport parameters and trains the CAT-Seg model for the segmentation task.  Experimental results on the MESS benchmark demonstrate that OV-COAST enhances the performance of CAT-Seg, surpassing other state-of-the-art models.